# Adaptive formation motion planning and control of autonomous underwater vehicles using deep reinforcement learning

Behnaz Hadi[1], Alireza Khosravi[*,1], Pouria Sarhadi[2]

[1] Department of Electrical and Computer Engineering, Babol Noshirvani University of Technology, Babol, Iran

[2] School of Physics, Engineering and Computer Science, University of Hertfordshire, Hatfield, United Kingdom

**Abstract**

Creating safe paths in unknown and uncertain environments is a challenging aspect of leader-follower formation control. In this architecture, the leader moves toward the target by taking optimal actions, and followers should also avoid obstacles while maintaining their desired formation shape. Most of the studies in this field have inspected formation control and obstacle avoidance separately. The present study proposes a new approach based on deep reinforcement learning (DRL) for end-to-end motion planning and control of under-actuated autonomous underwater vehicles (AUVs). The aim is to design optimal adaptive distributed controllers based on actor-critic structure for AUVs formation motion planning. This is accomplished by controlling the speed and heading of AUVs. In obstacle avoidance, two approaches have been deployed. In the first approach, the goal is to design control policies for the leader and followers such that each learns its own collision-free path. Moreover, the followers adhere to an overall formation maintenance policy. In the second approach, the leader solely learns the control policy, and safely leads the whole group towards the target. Here, the control policy of the followers is to maintain the predetermined distance and angle. In the presence of ocean currents, communication delays, and sensing errors, the robustness of the proposed method under realistically perturbed circumstances is shown. The efficiency of the algorithms has been evaluated and approved using a number of computer-based simulations.

## 1. Introduction

One of the important topics studied in the field of multiple underwater robots is formation control. Collaborative performance of complex tasks improves performance, reduces the cost and mission time, guarantees desired results, and increases the mission success rate compared to deploying individual vehicles [1, 2]. AUVs are unmanned autonomous vehicles that perform specific tasks in underwater environments for several hours to days using their control and navigation systems. AUVs can be deployed for investigating and mapping the seabed, exploration, identifying, and search and rescue missions, investigating geological threats and examining petroleum and gas pipelines, determining the water's biological parameters, research for navigation and control systems, performing hydrodynamic calculations, and so on [3-5]. The developments in applications of autonomous underwater vehicles indicate the significance of research in the field of improving their group movement control. A perfect, intelligent, and decision-making autonomous underwater vehicle is the aim of research in several studies in the field of underwater robots.

---

[*] Email: akhosravi@nit.ac.ir

Formation structure is a combination in which agents maintain the desired form and at the same time execute the assigned commands. Formation control has three main architectures: leader-follower, virtual structure, and behavioral, and each has its advantages and disadvantages. In the leader-follower approach, one or several AUVs are taken as leaders, and the other AUVs are designed as followers. The leaders will follow the trajectory or the designed reference path, but the followers will pursue the leader's status by maintaining the pre-determined desired values of distance and angle in order to generate and retain the shape of the formation. In the virtual structure, all agents hold a rigid geometric relationship based on a virtual point in order to generate the desired formation. In the behavioral architecture, a proper behavior is defined as a weighted combination of various control goals [6, 7]. Formation control has various applications, e.g., mobile robots [8], unmanned surface vehicles (USVs) [9], autonomous underwater vehicles (AUVs) [10], unmanned aerial vehicles (UAVs) [11], satellites [12], and spacecraft [13]. Among these mentioned fields, the formation control of AUVs is the most challenging due to uncertain and coupled system dynamics, environmental disturbances coming from ocean waves and currents, and underwater communications [14]. Among the methods that have been widely used in this field are adaptive control [15], sliding mode control [16, 17], predictive control [18], neural network-based control [19], graph theory [10], and so on. In [20], dynamic surface control along with a disturbance observer were used for time-variant formation control of AUVs. [21] used a lagrangian approach to control the formation of marine surface craft. The formation was accomplished through the application of constraint functions. Leader-follower formation control of under-actuated AUVs based on a combination of the Lyapunov theory and backstepping was presented in [22]. Bono et al. have reported formation control and obstacle avoidance of a group of AUVs using the distributed predictive control model [18]. A leader-follower architecture for multiple underwater vehicle manipulator systems (UVMS) was proposed by Heshmati et al. [23] to address the cooperative object transportation task, with a trajectory estimation scheme with the prescribed performance suggested for the followers. Evaluation of the majority of AUVs formation control techniques is presented in [6].

Despite the desirable control performance of classic control approaches, assumptions such as linearization of the system's dynamic model and precise identification of the model or the system's parameters are required, and most of these approaches demand complex analytical calculations, which make implementing them for actual applications on AUV systems challenging due to their highly nonlinear and uncertain coefficients in environments with unknown disturbances.

As one of the branches of machine learning, Reinforcement Learning (RL) has spread significantly in designing a class of optimal adaptive controllers with the actor-critic structure for non-linear and uncertain dynamic systems in both continuous and discrete times, which leads to an improvement in the AUV's level of autonomy [24]. In RL, one or more agents interact in real time with an environment that may be unknown to them, and based on the experiences they obtain, they learn to pick optimized strategies for achieving specific goals. Using just one efficiency scalar criterion, which is called the reinforcement or reward signal, this approach is able to train the agents in complex environments in an uncertain and stochastic manner without supervision [25]. A combination of RL and deep learning methods, called DRL, has become a powerful tool in solving complex problems of autonomous systems and adapting with unpredictable and uncertain conditions, which reduce the complexity of designing control systems while improving their efficiency [26].

Not much research has been done so far on using RL in AUVs formation control. The majority of the literature related to AUVs currently concentrates on using RL to control AUVs, path planning, and path following by single vehicles. In [27], DRL has been used to design an adaptive low-level controller for AUV. A combination of RL and imitation learning has been used in [28] to control a unmanned underwater vehicle (UUV). Sun et al. proposed using the RL method for the motion planning of an AUV. The proximal policy optimization (PPO) algorithm has been used to design surge and yaw control signals [29]. In [30], corrected Q-learning has been used to avoid obstacles in AUVs. As the AUV enters the pre-defined dangerous zone, the exploring process stops, and exploitation of control commands is performed until leaving the unsafe region. In [31], RL strategies have been used to control the docking of AUVs, and then the method is compared with classic control techniques. Zhang et al. have used interactive DRL in AUVs path-following. In this method, the deep Q-network (DQN) approach is combined with interactive RL, and the reward function includes both environmental and human rewards [32].

Due to the popularity of DRL, research has been done in the fields of cooperative and formation control for USVs, UAVs, mobile robots, and so on. Zhang et al. have proposed a circular formation control for fish-like robots [33]. In [34], formation control and obstacle avoidance based on RL and imitation learning have only been proposed for a couple of leader-follower with holonomic second-level dynamics. Decentralized asynchronous formation control and generalizable distributed formation for USVs based on the leader-follower architecture with the DRL approach have been performed in [35] and [36], respectively. Lin et al. have utilized a combination of DRL and a long short-term memory (LSTM) network to solve the problem of dynamic spectrum interaction regarding the formation of UAVs [37]. In [38], the formation control of a group of fixed-wing drones based on multi-objective RL is presented. In [39], the cooperative motion control of a flock of mobile robots is presented based on DRL with a prioritized experience replay buffer mechanism.

Motivated by the preceding discussion, a DRL-based distributed adaptive formation controller with obstacle avoidance and self-collision avoidance for under-actuated AUVs has been presented. Optimal adaptive controllers based on actor-critic architecture identify the performance of the current control policy and update the controller using such data to reduce formation motion planning errors. The model for the Markov decision process (MDP) process for end-to-end motion planning of the multi-AUVs formation has been presented. The system directly maps the state information of AUVs and the environment to control their heading and speed, realizing the end-to-end processing of the information. Obstacle avoidance throughout the exploration process is considered a challenging task for AUVs. Since the proposed approach is a model-free method, the dependence on the mathematical model has been eliminated, and an intelligent controller based on the behavior of the plant in an unknown environment is developed. Obstacle avoidance is done using two approaches. In the first approach, all the AUVs are equipped with obstacle avoidance modules. When the obstacles are within the detection range of each AUV, they are entered into its MDP model, and the change of direction and speed required to avoid the obstacles is done. In the second approach, only the leader is responsible for ensuring safety. When the obstacle appears and enters the leader's MDP model, considering the leader's awareness of the formation size, the path is chosen so that the whole group stays away from the obstacles.

The present paper is organized as follows. In Section 2, the fundamentals and formulation of the problem included: AUVs motion model, fundamentals of DRL, and formation control architecture are presented. The structure of DRL, including a definition of the state space, action space, and reward schemes, is introduced in Section 3. In Section 4, the details of the DRL algorithm for formation control and obstacle avoidance of multiple AUVs are presented. Simulations and experiments of various scenarios are presented in Section 5. Section 6 is devoted to conclusions.

## 2. Fundamentals and Formulation of the Problem

### 2.1. AUV Motion Model

This section aims to describe the kinematic and dynamic models of AUVs. To describe the motion of AUVs, two coordination systems are defined according to Fig. 1: the body-fixed frame {B}, and the earth-fixed frame {E}. The AUVs formation on a horizontal surface is dedicated to constant depth applications, such as investigating the seabed. The 3-DOF REMUS AUVs' dynamics and kinematics model with the nomenclature indicated in Table 1 can be described as follows [40]:

$$(m - X_{\dot{u}})\dot{u} - my_g\dot{r} - m(vr + x_g r^2) = X_{u|u|}|u|u + X_{vr}vr + X_{rr}r^2 + X_{prop}$$

$$(m - Y_{\dot{v}})\dot{v} + (mx_g - Y_{\dot{r}})\dot{r} + m(ur - y_g r^2) = Y_{v|v|}|v|v + Y_{r|r|}|r|r + Y_{ur}ur + Y_{uv}uv + Y_{uu\delta_r}u^2\delta_r$$

$$(I_z - N_{\dot{r}})\dot{r} + (mx_g - N_{\dot{v}})\dot{v} - my_g\dot{u} + m(x_g ur + y_g vr) = N_{v|v|}|v|v + N_{r|r|}|r|r + N_{ur}ur + N_{uv}uv + N_{uu\delta_r}u^2\delta_r \quad (1)$$

$$\dot{x} = \cos(\psi)u - \sin(\psi)v$$

$$\dot{y} = \sin(\psi)u + \cos(\psi)v$$

$$\dot{\psi} = r$$

where $[x, y, \psi]^T$ is the position vector of the AUV in the earth-fixed frame, which includes the $(x, y)$ coordinate and the yaw angle $\psi \in [0, 2\pi)$. The velocity vector of the AUV in the body-fixed frame is $[u, v, r]^T$, where $u$ is the surge velocity, $v$ is sway velocity, and $r$ is the yaw rate. $[X_{prop}, \delta_r]$ is the input vector, including the propeller thrust ($X_{prop}$) and deflection of the rudder angle ($\delta_r$).

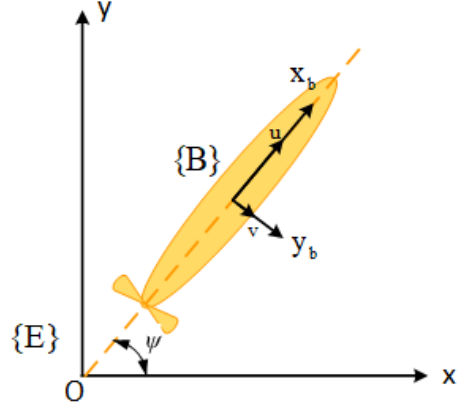

Fig. 1. AUV motion model

Table 1. Definition of AUV coefficients

| Parameter | Unit | Description | Parameter | Unit | Description |
|---|---|---|---|---|---|
| $X_{u\|u\|}$ | $kg/m$ | | $Y_{ur}$ | $kg/rad$ | Added mass cross term and fin lift |
| $N_{v\|v\|}$ | $kg$ | | $N_{ur}$ | $kg.m/rad$ | |
| $Y_{v\|v\|}$ | $kg/m$ | Cross flow drag | $N_{uv}$ | $kg$ | Body and fin lift and Munk moment |
| $Y_{r\|r\|}$ | $kg.m/rad^2$ | | $Y_{uv}$ | $kg/m$ | Body lift force and fin lift |
| $N_{r\|r\|}$ | $kg.m^2/rad^2$ | | $Y_{uu\delta_r}$ | $kg/(m.rad)$ | Fin lift force |
| $X_{\dot{u}}$ | $kg$ | | $N_{uu\delta_r}$ | $kg/rad$ | Fin lift moment |
| $N_{\dot{v}}$ | $kg.m$ | Added mass | $m$ | $kg$ | Vehicle mass |
| $N_{\dot{r}}$ | $kg.m^2/rad$ | | $I_z$ | $kg.m^2$ | Vehicle moment of inertia around z axis |
| $X_{vr}$ | $kg/rad$ | Added mass cross-term | $x_g, y_g$ | $M$ | Center of gravity in x and y directions with respect to the center of buoyancy |
| $X_{rr}$ | $kg.m/rad$ | | | | |

## 2.2 DRL method

Reinforcement learning (RL) is a decision-making framework in which an agent learns the desired behavior or policy by interacting directly with the environment (Fig. 2). At every time step, the agent performs action *a* in the state of *s*. Hence, it is now in the new state of *s'* and receives the reward *r* from the environment. A Markov decision-making chain is employed for modeling the reinforcement environment, which is expressed in a 4-element tuple $[s, a, r, p]$. Here, *s* is the dynamic behavior of the environment, *a* is the action performed by the agent, *r* is the reward from the environment, and *p* is the transition probability function. After a long interaction with the environment, the agent learns an optimal policy that maximizes the overall expected return. The expected return is defined as the discounted sum of future rewards as follows:

$$R_t = r_{t+1} + \gamma r_{t+2} + \gamma^2 r_{t+3} + \cdots = \sum_{k=0}^{\infty} \gamma^k r_{t+(k+1)} \tag{2}$$

where $\gamma \in [0\ 1]$ is the discount factor that determines the current value of future rewards. The discounted expected state-value function defined as following, starting from the state $s$ and following the $\pi$ policy:

$$V^\pi(s) = E_\pi[R_t | s_t = s] = E_\pi[\sum_{k=0}^{\infty} \gamma^k r_{t+(k+1)} | s_t = s] \tag{3}$$

Similarly, the action-value function, which expresses the value of the action $a$ in the state of $s$ under the $\pi$ policy is defined as following:

$$Q^\pi(s,a) = E_\pi[R_t | s_t = s, a_t = a] = E_\pi[\sum_{k=0}^{\infty} \gamma^k r_{t+(k+1)} | s_t = s, a_t = a] \tag{4}$$

The optimal state-value function yields the maximum discount reward when an agent starts from the state s and executes the optimal policy.

$$V^*(s) = \max_a E[r_{t+1} + \gamma V^*(s_{t+1}) | s_t = s, a_t = a] \tag{5}$$

The equation mentioned is called the Bellman Optimality Equation, which states the value of a state under an optimal policy should be equal to the expected return for the best action in that state. Similarly, Bellman's equation for the action-value function is as follows:

$$Q^*(s,a) = E[r_{t+1} + \gamma \max_a Q^*(s_{t+1}, a') | s_t = s, a_t = a] \tag{6}$$

where $V^*(s) = \max_a Q^*(s,a)$ is for all s. When $Q^*$ is identified through interactions, the optimal policy can be obtained directly from the following equation:

$$\pi^*(s) = \arg\max_a Q^*(s,a) \tag{7}$$

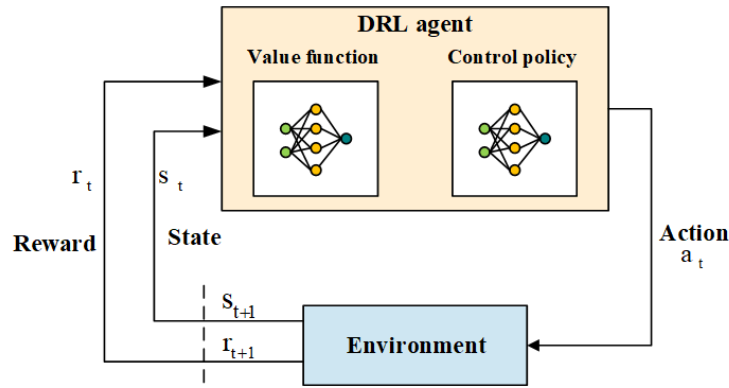

Fig. 2. Interaction of the deep agent with the environment in the Markov decision-making process

The standard RL approaches are based on discrete state and action spaces. Thus, the amounts of state value or action value are recorded in tables. One major advantage of these approaches is that the precise optimized value function and the optimized policy can always be found [25]. The RL algorithms based on tables suffer from the curse of dimensionality in problems with large or continuous spaces. To solve this problem, a combination of function approximators like deep neural networks is used. Deep neural networks are structures of sequential layers that transform a high-dimensional input into a reduced output feature, and the parameters of the network are trained using a supervised approach that utilizes training algorithms such as the gradient descend method in backpropagation algorithms. Therefore, there is no need to keep a table containing the data for the state-action value function, and consequently, vast memory and precise information on the environment will not be necessary. Despite solving problems with high-dimensional observation space by the DQN algorithm can work with small and discrete action spaces [41]. Since most of the physical control tasks have continuous action spaces and high dimensions, the deep deterministic policy gradient (DDPG) with the actor-critic structure was introduced, which can learn the policies in high-dimensional continuous action spaces [42]. The DDPG algorithm uses two deep neural networks to approximate the state-action value function and policy, along with replay buffer ideas and target networks. The replay buffer is used to break the nature of the Markov among the sampled data, and two target networks are used to guarantee the stability of the training process. Despite the desirable performance of the DDPG algorithm, it doesn't have acceptable performance in complex environments due to overfitting of the current policy, hyper-parameters, and other settings. That's why the twin delayed deep deterministic policy (TD3) algorithm was presented, which enhances the learning speed and improves the performance through the use of double-critic networks, smoothing the target policy, and policy update delay [43].

*2.2.1 The TD3 Algorithm*

The TD3 algorithm is an online RL algorithm which is expressed based on the DDPG algorithm. This approach executes a method similar to the double-DQN, which reduces over fittings in function approximation and causes a delay in the update frequency of the actor network, and eliminates sensitivity and instability in the DDPG network by adding noise to the target actor network [43]. The block diagram of the algorithm's structure is depicted in Fig.3.

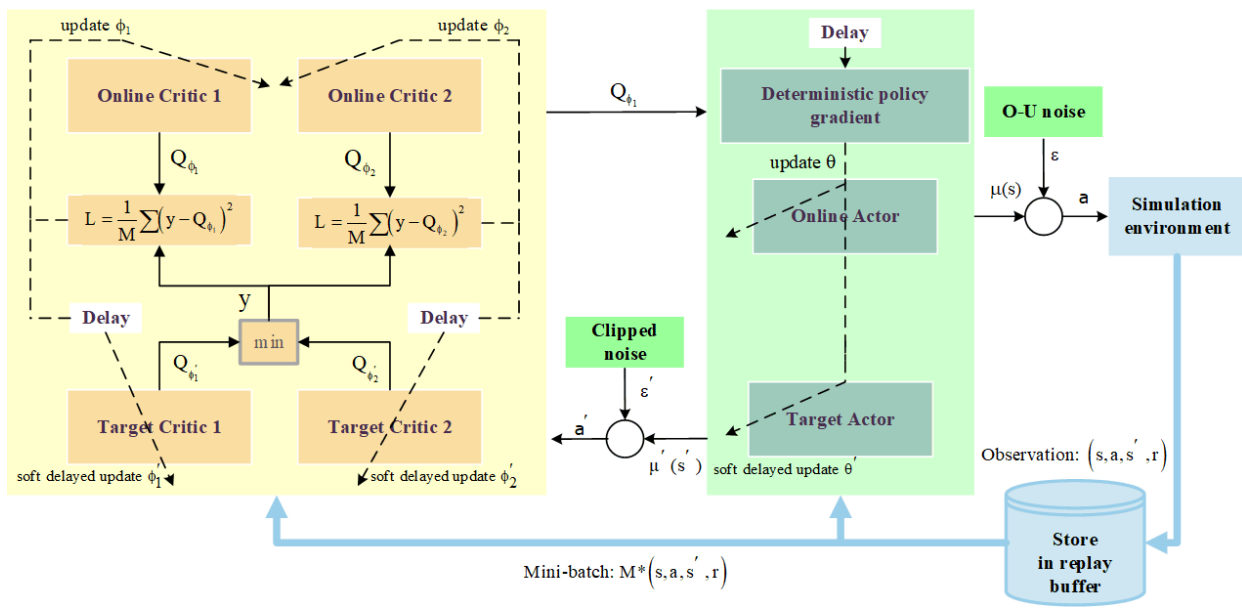

Fig. 3. The TD3 framework based on actor-critic structure

According to this figure, the algorithm is based on an actor-critic architecture and has double-critic networks. The state transitions required for updating the networks are stored in the buffer by selecting an action and applying it to the environment. When a certain number of transitions are stored in the buffer, a mini-batch of the buffered transition states is sampled to find the policy. The critic network calculates the state-action value function of the online critic network using these buffer samples and by minimizing the mean squares of the Bellman error based on the target value function of the smaller critic using the gradient descent method, and hence the networks' parameters are updated. Then, the actor network updates the policy network's parameters with a delay using the calculated value function based on the deterministic gradient theorem for the expected return optimization. This procedure is repeated until achieving the optimal policy for the environment.

*2.3 Formation Control and Obstacle Avoidance Architecture of Multi-AUVs*

According to Fig. 4, AUVs in the leader-follower formation introduced in the present paper can be divided into three groups, which include: the leader AUV, which does not follow any other AUV and tries to learn a policy that helps them reach their destination in the shortest probable distance and avoid possible obstacles on the way or other AUVs; and the followers on the left or right of the leader, which are supposed to find an optimal policy that helps them reach a predetermined distance and angle with the leader and maintain these values throughout the mission; if obstacles show up on their path, they avoid them, and they also avoid colliding with each other. Each follower can act as the leader for the other AUVs that are added to the formation to increase their number. AUVs could be outfitted with a variety of onboard sensors. A depth sensor, an Inertial Measurement Unit (IMU), compasses, a Doppler velocity log (DVL), a sonar sensor, a communication module, and a Global Positioning System (GPS) for on-surface corrections could be considered. The IMU can measure azimuth, triaxial angular velocity, and triaxial acceleration. A compass indicates directions to the cardinal geographic directions. A DVL is used to determine underwater speeds. The sonar sensor utilizes acoustics to detect obstacles. To establish formation control, AUVs should exchange essential information via wireless communication. The follower needs to track the leader. As a result, they do not require the leader's sophisticated sensors. This article discusses the challenges of controlling the leader-follower formation of multiple AUVs. We proposed a novel formation control algorithm with both a unidirectional flow of information for the first motion planning method and a bidirectional flow for the second one. In the first method, followers obtain information from the leader's AUV but not vice versa. To avoid obstacles, all AUVs are outfitted with forward-looking sonar. The leader AUV will continuously transmit its real-time states to the two follower AUVs to maintain the desired distances and orientations to the leader. In the second method, the leader and follower AUVs exchange information with each other. Only the leader has sonar sensors to detect obstacles, and the followers must only follow the leader.

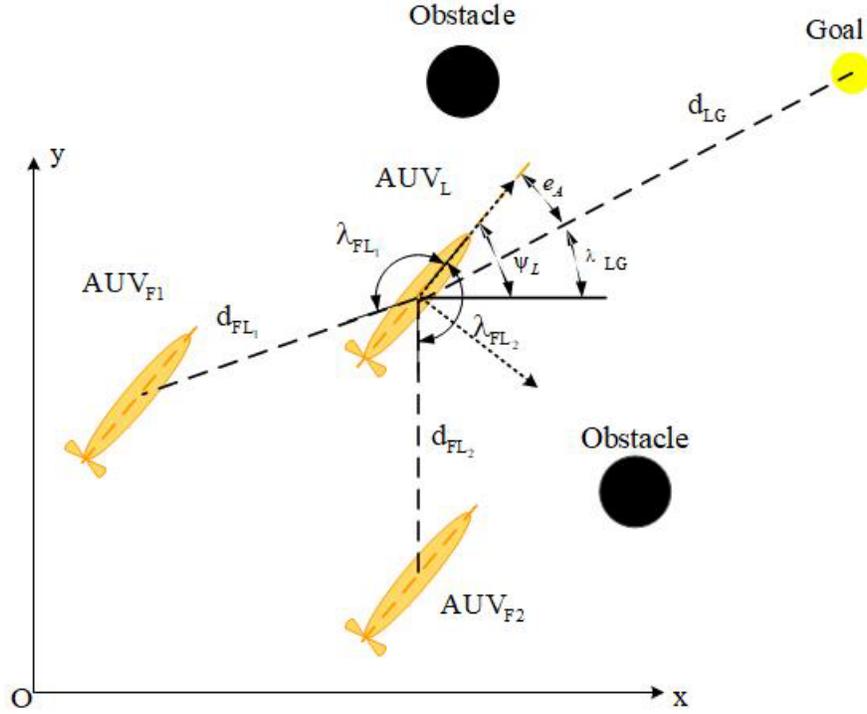

Fig. 4. Leader-Follower architecture for the multi-AUVs formation control

## 3. The Structure of DRL for Formation motion Planning of multi-AUVs

Fig. 5 illustrates the distributed formation model based on TD3 for the formation control of under-actuated AUVs. The training model of formation control is a repeated process of interacting with the environment. Based on observing the state of the environment $(s_i)$, each AUV can make reasonable decisions. By combining task objectives and state observation, the action selection can be evaluated. According to the assessed value, formation motion planning of multiple AUVs can be trained. Eventually, the model executes the selected action in the corresponding updated environment and recaptures the observation state. The mentioned procedure is repeated until the formation motion planning model selects the expected action. This way, the model can learn how to make ideal control decisions for the formation control of multiple AUVs. According to Fig. 5, the outputs of the decision-making network $(a_i)$ are the actions of individual AUVs. Each AUV is transferred into a new state after executing the action. Then a reward $(r_i)$ is generated by measuring the behavior of the AUV. Finally, the data for any AUV is stored in the corresponding experience pool, and then the experience data can be picked for shaping the training data collection. The experience pool is used for storing the records of the navigation experience of each AUV. The state data in a single execution, the actions of the decision-making network, and the corresponding rewards form an experience record altogether. Every AUV can be trained to plan formation paths and avoid obstacles in the decision-making network.

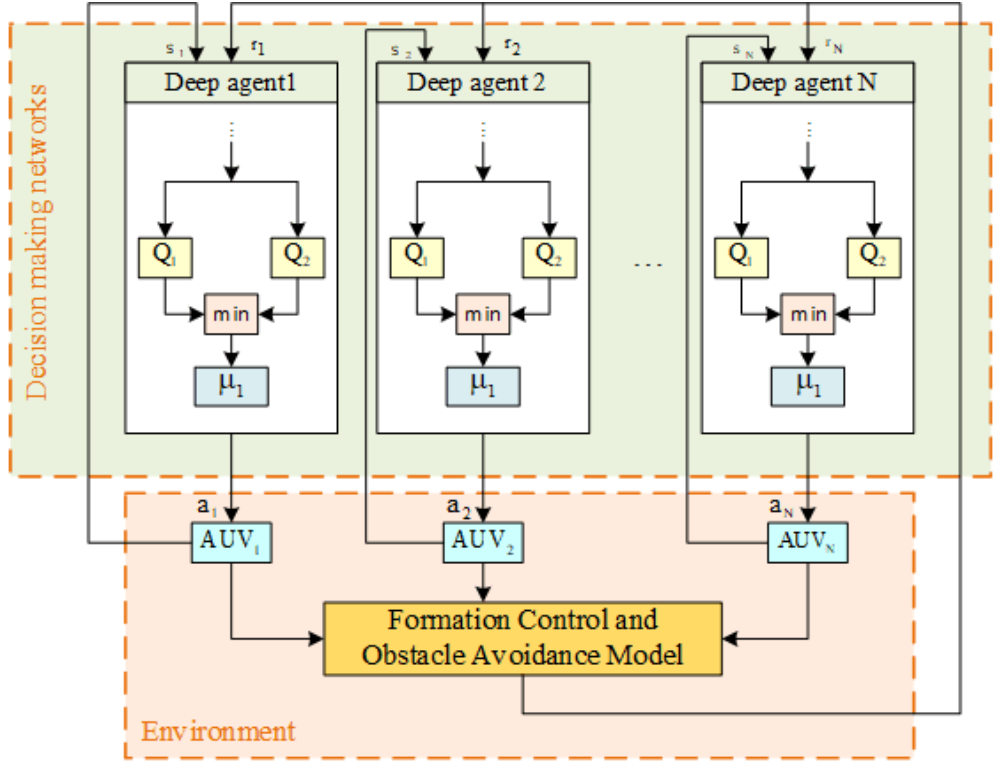

Fig. 5. Distributed DRL structure for formation control and obstacle avoidance multi-AUVs

In obstacle avoidance, two approaches are applied. In the first approach, obstacle avoidance decisions are made by each AUV present in the formation. According to the designed reward function, the follower AUVs learn to avoid obstacles while maintaining the formation. When facing an obstacle, avoiding a collision with it has a higher priority than formation maintenance. In the second approach, making decisions regarding obstacles is solely the leader's responsibility. The leader identifies the obstacle, and, being aware of the size of the formation, learns a policy appropriate for keeping the whole group away from it. According to the two mentioned approaches, various MDPs are considered when facing obstacles.

*3.1 Setting the State Space and the Action Space*

To execute the formation control and obstacle avoidance of under-actuated AUVs based on the leader-follower architecture, first the state space and the action space should be introduced. A MDP based on the leader-follower architecture is defined for each AUV. According to the role of each AUV in the formation architecture, the corresponding state space is designed. The obstacles are identified using the AUVs' onboard sensors. According to the range of the AUVs' sensors, the distance from the identified obstacles is added to the MDP of each agent.

*3.1.1 Designing the State Space for the Leader*

The aim of the leader is to reach the destination safely by using the shortest possible path. Thus, the state space is based on the leader's direction error from the target, the distance from the identified obstacles, and the velocity vector.

$$S_{Leader} = \left[ e_A, v^T, \frac{d_{LO} - r_{det}}{r_{det}} \right], \tag{8}$$

$$\frac{d_{LO} - r_{det}}{r_{det}} = \begin{cases} \text{value} & \text{if obstacles are in AUV's sensor range detection} \\ 0 & \text{else} \end{cases}$$

where $e_A = \lambda_{LG} - \psi_L$ is the direction error of the leader toward the target, and $\lambda_{LG}$ is the target angle relative to the x-axis (Fig. 4). $\psi_L$ is the heading angle of the leader AUV. The AUV's velocity vector is $v^T = [u, v, r]_{1 \times 3}$. $d_{LO}$ is the $N \times 1$ vector of the distance from identified obstacles, and $N$ is the number of the obstacles. $r_{det}$ is the maximum detection range of the AUV's sensor. The state space for both obstacle avoidance approaches is identical for the leader.

### 3.1.2 Designing the State Space for the followers

The followers intend to learn a policy that helps them maintain a predetermined distance and angle towards the leader while avoiding obstacles when facing them. For this purpose, the follower state space is defined as follows:

$$S_{Follower} = \left[ e_d, e_A, \psi_L, \psi_F, \frac{V_L - V_d}{V_d}, \frac{V_F - V_d}{V_d}, \frac{d_{LO} - r_{det}}{r_{det}} \right] \tag{9}$$

$$\frac{d_{LO} - r_{det}}{r_{det}} = \begin{cases} \text{value} & \text{if obstacles are in AUV's sensor range detection} \\ 0 & \text{else} \end{cases}$$

where $e_d = \frac{d_{FL} - d_{desired}}{d_{desired}}$ is the distance scaled error, $d_{FL}$ and $d_{desired}$ are the distance between the leader and follower and the desired distance respectively. $e_A = \lambda_{FL} - \lambda_{desired}$ is the formation's angle error, where $\lambda_{FL}$ is the angle between the leader-follower's connecting line and the leader's movement direction, and $\lambda_{desired}$ is the formation desired angle. $\psi_L$ and $\psi_F$ are the heading angles of leader and follower respectively. $V_i = \sqrt{u^2 + v^2}$, $i = L, F$ is resultant velocity and $V_d$ is a desired formation velocity. $d_{LO}$ is the $N \times 1$ vector of the distance from identified obstacles, and $N$ is the number of obstacles. $r_{det}$ is the maximum detection range of the AUV's sensor. In the second approach to obstacle avoidance in the state space the element of obstacle is eliminated from the MDP, since obstacle identification is exclusively performed by the leader.

*3.1.4 The Action Space of the Leader and Followers*

As mentioned earlier, the leader intends to arrive at the destination safely and for the followers to maintain their formation according to the leader and avoid obstacles throughout navigation, which requires velocity and direction control. Controlling the propeller thrust $X_{prop}$ and rudder angle $\delta_r$ are used to accomplish this.

$$A = \left[ X_{prop}, \delta_r \right] \tag{10}$$

*3.2 Shaping the Rewards of AUVs' Formation*

Reward functions for the AUVs' formation and obstacle avoidance are designed based on two different approaches. The rewards' function is basically to encourage expected behaviors and punish unexpected ones. Three categories of reward functions are defined for the leader and followers. The leader's reward functions are considered independently to find an optimal path towards the target while avoiding obstacles and collisions with other AUVs. For followers, rewards are defined to shape and maintain formation, avoid obstacles, and collide with the other AUVs. The part about obstacle avoidance is removed from the followers' reward function in the second obstacle avoidance approach.

*3.2.1 Designing the Leader's Reward*

1) target reward:

This reward guarantees that the leader is able to reach the destination.

$$r_A = -\left| \lambda_{LG} - \psi_L \right| \tag{11}$$

where $\lambda_{LG}$ is the line of sight angle target with the leader, and $\psi_L$ is the leader's heading angle.

2) Obstacle avoidance reward:

**Approach 1:** In this approach, each AUV learns a policy to avoid obstacles. According to Fig. 6a, first, a circular shell is made for the AUV model. A shell overlapping with an obstacle indicates a collision. The scaled distance of all obstacles inside sensors' detection range is entered into the corresponding AUV's MDP model in Section 3.1.1, and thus, the reward function is defined as follows:

$$r_{OA1} = \sum_{i=1}^{\ell} r_i, \qquad r_i = \begin{cases} 0 & \text{if } d_{AO} > d_{avoid} \\ -\left| d_{avoid} - d_{AO} \right| & \text{otherwise} \end{cases} \tag{12}$$

where $\ell$ denotes the number of obstacles sensed within the detection range of the AUV's sensors. $d_{AO}$ is the distance between AUV and obstacle and $d_{avoid} = r_A + d_{safe} + r_{ob}$, which $r_A$ is AUV's shell radius, $d_{safe}$ is the safe distance width from the shell of AUV, and $r_{ob}$ is the obstacle radius.

If the distance of the AUV from the obstacles is more than the $d_{avoid}$, it receives zero. Otherwise, it is punished.

**Approach 2:** In this approach, formation motion planning is exclusively performed by the leader. The followers' duty is to maintain the desired distance and angle from the leader. According to Fig. 6b, a triangle whose vertices are the positions of the leader and left and right followers is considered. When an obstacle enters the leader's detection range, the center of the triangle's circumscribed circle is calculated based on the location of the leader and followers. Using the distance from the center of the circumscribed circle to the obstacle, the reward function is defined as follows:

$$r_{OA2} = \sum_{i=1}^{\ell} r_i, \qquad r_i = \begin{cases} 0 & \text{if } d_{CO} > d_{avoid} \\ -|d_{avoid} - d_{CO}| & \text{otherwise} \end{cases} \qquad (13)$$

where $d_{CO}$ is the distance from the center of the circumscribed circle to the obstacle, and $d_{avoid} = r_{cir} + d_{safe} + r_{ob}$, which $r_{cir}$ is the radius of the circumscribed circle, $d_{safe}$ is the safe distance width from center of gravity of AUV, and $r_{ob}$ is the obstacle radius.

**Remark:** In the first approach, all AUVs are equipped with sonar detection sensors to assist them in making the correct decision if they come into contact with something. In addition to maintaining the desired distance and direction, the followers must avoid obstacles and return to the desired formation after traversing the danger zone. Each AUV is equipped with an obstacle avoidance module. Thus, it can be a safer strategy. But in the second method, only the AUV leader is equipped with obstacle-detecting sensors, while the followers maintain the desired distance and orientation. When the leader detects an obstacle, if there is a chance of the group colliding with it based on the position of the followers, the leader alters the group's movement path and attempts to move the entire group out of the danger zone. In both methods, the followers require the leader's state to produce the proper formation. The necessary data is communicated to the followers using low-cost wireless modems mounted aboard the AUVs. In the first method, the leader does not need to be aware of the follower's state. Thus, information is transmitted in a unidirectional way. In the second strategy, the leader additionally requires information about the position of the followers. They are exchanging data in this bidirectional way. This method is also a better choice for the centralized formation framework. Overall, those two methods are similar, but they take different approaches to implementation, and one could be chosen based on the system design. The simulation section demonstrates that both proposed obstacle avoidance methods work properly.

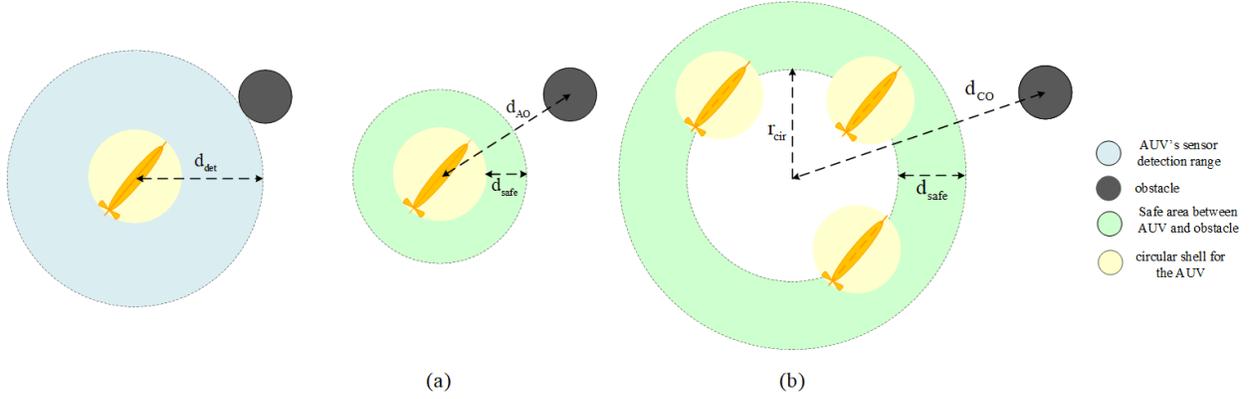

Fig. 6. Two obstacle avoidance approaches: (a) Approach1: Each AUV senses the obstacles in the range of sensors detection, it can obtain the relative distance to the obstacle. (b) Approach2: Only leader senses the obstacles

3) Self-collision avoidance reward:

$$r_{CA} = \begin{cases} 0 & \text{if } d_{AA} > d_{avoid} \\ -|d_{avoid} - d_{AA}| & \text{otherwise} \end{cases} \quad (14)$$

where, $d_{AA}$ is the distance between two AUVs (Fig. 7). This reward is considered for the AUVs not colliding with each other. $d_{avoid} = 2(r_A + d_{safe})$.

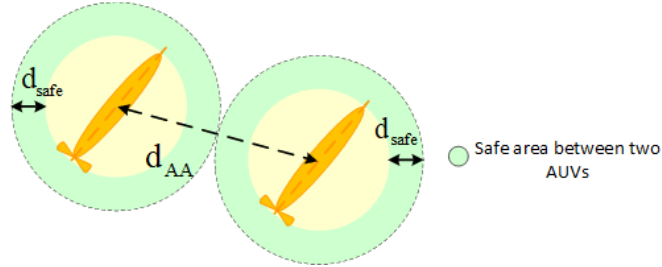

**Fig. 7.** Collision avoidance: If the distance between the two AUVs is less than the safe distance, they receive a negative reward

4) Reducing control effort:

One of the practical challenges in the design of autonomous vehicle controllers is the saturation of actuators [44]. Thus, to account for practical constraints, the following term is used to reduce overall control effort:

$$r_E = [-|X_{prop}|, \ -|\delta_r|] \quad (15)$$

where $X_{prop}$ and $\delta_r$ are the propeller thrust and rudder angle, respectively.

5) Total reward:

The total reward function is defined as the weighted sum of all the mentioned reward functions:

$$r_L = w_A r_A + w_{OA} r_{OA} + w_{CA} r_{CA} + [w_{E1}, \ w_{E2}] r_E^T \tag{16}$$

where $w_A$, $w_{OA}$, $w_{CA}$, $w_{E1}$ and $w_{E2}$ are positive constants. Each of the above gains indicate the importance of each section in the control policy.

### 3.2.2 Designing the Follower's Reward function

In followers, the aim is to learn a policy that can maintain the desired distance and angle towards the leader while tracking it and avoiding obstacles and other AUVs. The rewards for avoiding obstacles, avoiding collisions with other AUVs, and limiting the amplitude of the control signals are calculated using the Eq. (13), Eq. (15) and Eq. (16) in Section 3.2.1. As a result, the formation reward functions are as follows:

1) Formation distance reward:

$$r_{Fd} = -\left|\frac{d_{FL} - d_{desired}}{d_{desired}}\right| \tag{17}$$

$d_{FL}$ is the distance between the follower and the corresponding leader. The desired formation distance is expressed as $d_{desired}$.

2) Formation angle reward:

$$r_{FA} = -\left|\lambda_{FL} - \lambda_{desired}\right| \tag{18}$$

$\lambda_{FL}$ is the formation angle (the angle between the line connecting the leader and the follower and the leader's direction) (Fig. 5) and $\lambda_{desired}$ is the desired formation angle.

3) Total reward:

The total reward function, is defined as the weighted sum of all the reward functions:

$$r_F = w_{Fd} r_{Fd} + w_{FA} r_{FA} + w_{OA} r_{OA} + w_{CA} r_{CA} + [w_{E1}, \ w_{E2}] r_E^T \tag{19}$$

where $w_{Fd}$, $w_{FA}$, $w_{OA}$, $w_{CA}$, $w_{E1}$ and $w_{E2}$ are positive constants. Each of the above weights indicate the importance of each section in the control policy. In the second approach of obstacle avoidance, the term $r_{OA}$ is removed.

The observations are collected according to the 3.1.1 and 3.1.2 sections. Then, the action selected based on the suggested formation control model for generating control commands for the AUVs' actuators is executed. Accordingly, target reaching error of the leader and the formation error for the followers towards achieving desired control targets are evaluated. These steps are repeated until the AUVs safely reach their destination in the desired formation. In the control cycle, AUVs perform the environmental awareness, decision-making, and execution processes in that order. With environmental awareness, the status of AUVs in the formation can be observed, and the situation of the AUVs in the formation is then updated accordingly.

## 4. DRL Algorithm for Formation Control and Obstacle Avoidance of Multiple AUVs

The proposed formation control and obstacle avoidance model seeks an optimal policy for each agent using the TD3 algorithm. Before the start of the episodes, all the settings and formation configuration are done, including determining the leader and follower AUVs, desired angles and distances, the dimensions of the training area, the training rate, the exploration noise for the actor network, the smoothing noise for the target network, etc. In each episode, the variables of the velocity vector and the AUVs' position are reset. A random target zone is selected, and the obstacles are distributed in the training zone randomly. In the algorithm, the critic has four networks, of which two have similar structures, i.e., online networks with $\phi_1$ and $\phi_2$ parameters, and target networks with $\phi_1'$ and $\phi_2'$ parameters. The actor has two online and target networks with similar structures, with $\theta$ and $\theta'$ parameters, respectively. In order to have proper exploration in state and action spaces while choosing the policy, the noise $N_t$ is added as follows:

$$a_t = \mu_\theta(s_t) + N_t \tag{20}$$

In each time step, by applying action to AUV, new states $s_{t+1}$ based on Eq. (9) are obtained, and the instant reward $r_t$ based on the reward function designed for control purposes is received and stored in the experience buffer (D) along with the applied action and the system's current status. The training process of the online actor network is updated according to the random data sampled from the buffer, $M*(s_i, a_i, r_i, s_{i+1})$ according to the following equation based on a deterministic policy gradient with a lower frequency than the critic networks. By doing so, the critic network becomes more stable before being used for training the target network, and the errors are reduced.

$$\nabla_\theta J(\theta) = \frac{1}{M} \sum_{s \in M} \nabla_\theta Q_{\phi_1}(s, \mu_\theta(s)) \tag{21}$$

where M is the size of the mini-batch. $Q_{\phi_1}$ is the value function of the online critic network with the parameter of $\phi_1$. The $\phi_1$ parameter is updated through minimizing the mean square of Bellman error which is based on the smaller target value function.

$$L(\phi_k) = \frac{1}{M} \sum_{i=1}^{M} (y_i - Q(s_i, a_i | \phi_k))^2 \quad , k = 1, 2 \tag{22}$$

where $y_i$ is the value of the target state-action with the smaller value generated by the target deep networks $Q_{\phi_1'}$ and $Q_{\phi_2'}$, which are parametrized by $\phi_1'$ and $\phi_2'$ including:

$$y_i = r + \gamma \min_{j=1,2} Q_{\phi_j'}(s_{i+1}, a'(s_{i+1})) \tag{23}$$

Applying the smoothing regularization strategy to the target policy and adding clipped noise to the deterministic output of the target actor network $a'$ for the state $s_{i+1}$, prevent high variance when updating the critic network. The online critic network's parameters $\phi_1$ and $\phi_2$ are updated using the stochastic gradient descent approach. With a frequency equal to the update frequency of the actor's policy function, the parameters of the target actor network $\theta'$ and the target critic networks $\phi_1'$ and $\phi_2'$ are soft updated with the aim of enhancing the stability of the learning process. This procedure is repeated in every episode until the leader AUV reaches the destination or the length of the episode's step is over. Consequently, by using the TD3 algorithm, the formation motion planning of multi-AUVs is presented in Algorithm 1. In order to estimate the value function (state-action) and the policy for actor and critic networks of the DRL algorithm, an architecture with two hidden layers has been utilized.

## 5. Simulation results

To verify the efficiency of the proposed method, the motion planning and formation control of under-actuated AUVs are assessed in various elaborated scenarios. To this end, formation keeping in waypoint tracking and in the presence of obstacles are considered. Furthermore, the performance and robustness of the algorithms are evaluated in the presence of ocean currents, variable communication delay and navigation errors. The obtained results are discussed in this Section.

### *5.1. Simulation Setting*

#### *5.1.1. Simulation environment setting*

An AMD Ryzen 7 3800XT 8-Core, 3.89 GHz CPU processor and an NVIDIA GeForce RTX 2060 GPU were used to run algorithm 1 for the 3-DOF REMUS model AUV in MATLAB. The values of the model's parameters are shown in Table 2. The training zone is a 500 × 500 m² square. The AUVs are placed in the central area of this square. The target is a circle with a radius of 3 meters at the boundary of the training zone. Each AUV with a length of 1-meter is inscribed in a circle with a 0.5-meter radius. The parameters of the dynamic model used in simulations $AUV_i$ ($i = 1,2,3$) are adapted from [40]. The authorized area of control signals is in the range of $X_{prop} \in [3,13]$ (N) and $\delta_r \in [-20, 20]$ (deg). The desired formation velocity is 1.5 m/s. The obstacles are distributed randomly in a 14-meter radius inside the training zone.

#### *5.1.2. ML algorithm training and parameter setting*

The structure of the actor-critic networks is similar, which means that double-layer, fully connected networks of 400 and 300 neurons with a RELU activator function have been used. These values have been utilized based on the dimensions of state and action spaces, which is satisfactory to approximate the state-action and policy functions in this research. Using the Ornstein-Uhlenbeck process noise, action is chosen to fully exploit the state and action spaces. The noise process is defined as follows:

$$N_{k+1} = N_k + \alpha_{OU}(\mu_{OU} - N_k) + \sigma_{OU} N_G(0,1) \tag{24}$$

where $N_k$ and $N_{k+1}$ are the values of the Ornstein-Uhlenbeck process at times k and k+1, respectively. $N_G(0,1)$ is a Gaussian noise with a zero mean and a 1 standard deviation. $\alpha_{OU}$, $\mu_{OU}$, and $\sigma_{OU}$ are the parameters of the Ornstein-Uhlenbeck process. $\alpha_{OU}$ is a constant that determines how quickly the noise output is attracted to the mean. $\mu_{OU}$ and $\sigma_{OU}$ are the mean and variance of the noise model, respectively. The values of the TD3 parameters based on formation motion planning and AUV control are listed in Table 3.

Table 2. The values of the REMUS AUV coefficients [40]

| Parameters | Values | Unit | Parameters | Values | Unit |
|---|---|---|---|---|---|
| $X_{u\|u\|}$ | -1.62e+000 | $kg/m$ | $X_{rr}$ | -1.93e+000 | $kg.m/rad$ |
| $N_{v\|v\|}$ | -3.18e+000 | $kg$ | $Y_{ur}$ | 5.22e+000 | $kg/rad$ |
| $Y_{v\|v\|}$ | -1.31e+000 | $kg/m$ | $N_{ur}$ | -2.00e+000 | $kg.m/rad$ |
| $Y_{r\|r\|}$ | 6.32e-001 | $kg.m/rad^2$ | $N_{uv}$ | -2.40e+001 | $kg$ |
| $N_{r\|r\|}$ | -9.40e+001 | $kg.m^2/rad^2$ | $Y_{uv}$ | -2.86e+001 | $kg/m$ |
| $X_{\dot{u}}$ | -9.40e-001 | $kg$ | $Y_{uu\delta_r}$ | 9.64e+000 | $kg/(m.rad)$ |
| $Y_{\dot{v}}$ | -3.55e+001 | $kg$ | $N_{uu\delta_r}$ | -6.15e+000 | $kg/rad$ |
| $N_{\dot{v}}$ | 1.93e+000 | $kg.m$ | $m$ | 30.51e+000 | $kg$ |
| $N_{\dot{r}}$ | -4.88e+000 | $kg.m^2/rad$ | $I_z$ | 3.45e+000 | $kg.m^2$ |
| $X_{vr}$ | 3.55e+001 | $kg/rad$ | $x_g, y_g$ | 0.0e+000 | $M$ |

The AUV can move 300 steps in each episode. The AUV's action is the surge force and the moment of yaw, so the heading and position of AUVs are updated by Eq. (1) and Eq. (2). Termination occurs when the leader AUV reaches the target area or the entire time step of each episode has passed.

Table 3: networks' parameters

| Parameters | Value |
|---|---|
| Actor network learning rate | 0.001 |
| Critic network learning rate | 0.0001 |
| Discount factor | 0.99 |
| Memory size | 1e6 |
| Smooth update | 0.005 |
| Policy and target delay update | 2 |
| Target policy noise variance | 0.2 |
| Exploration variance | 0.1 |
| Sample time | 0.1 |
| $\mu_{OU}$ | 0 |
| $\alpha_{OU}$ | 0.15 |

| Algorithm 1: Adaptive formation control and obstacle avoidance based on DRL |
|---|

1. Configure desired formation, initial poses $[x_0, y_0, \psi_0]$, number of episodes T, number of steps in each episode N, reply buffer size $D$, mini-batch size M, the learning rate for decision-making networks $\gamma$
2. Initialization online critic networks $Q_{\phi_1}$, $Q_{\phi_2}$ and the online actor $\mu_\theta$
3. Initialization target critic networks $Q_{\phi_1'}$, $Q_{\phi_2'}$ and target actor network $\mu_{\theta'}'$
4. For episode = 1, T do
5. Initialize a random process $(N_t)$ for action exploration
6. Receive initial observation states (Eq. (8) and Eq. (9))
7. For t = 1, N do
8. Selecting the action $a_t = \mu(s|\theta) + N_t$ according to the current policy and the explored noise
9. Executing the action $a_t$ on AUV and observing the new state $s_{t+1}$ (Eq. (1)) and obtaining the reward $r_t$ (Eq. (16) and Eq. (19))
10. Storing the transfer $(s_t, a_t, r_t, s_{t+1})$ in the experience reply buffer $D$
11. Sampling a random mini-batch ($M * (s_i, a_i, r_i, s_{i+1})$) from the experience reply buffer $D$
12. For i =1, M do
13. Calculation $a_{i+1}' = \mu'(s_{i+1}|\theta') +$ clipped noise
14. Calculating the value of target state-action using the equation $y_i = r_i + \gamma \min_{j=1,2} Q_{\phi_j'}(s_{i+1}, a_{i+1}'|\phi_j')$
15. End for
16. Updating the critic networks by minimizing the cost function of Bellman equation errors

$$\phi_1 = r_i + \arg\min_{\phi_1} \frac{1}{M} \sum_{i=1}^{M} (y_i - Q_{\phi_1}(s_i, a_i|\phi_1))$$

$$\phi_2 = r_i + \arg\min_{\phi_2} \frac{1}{M} \sum_{i=1}^{M} (y_i - Q_{\phi_1}(s_i, a_i|\phi_2))$$

17. If t mod d then
18. Updating the online actor policy (θ) with a frequency lower than updating the critic networks, using the sampled deterministic policy gradient $\nabla_\theta J(\theta) = \frac{1}{M} \sum_{s \in M} \nabla_\theta Q_{\phi_1}(s, \mu_\theta(s))$,

soft updating the target networks
$$\phi_i' = \tau \phi_i' + (1-\tau) \phi_i', \text{ for } i = 1, 2$$
$$\theta' = \tau \theta' + (1-\tau) \theta'$$

19. End if
20. End for
21. End for

## 5.2. Motion planning and formation control: Approach 1

Three AUVs are considered for a triangular formation. The leader's initial position ($AUV_L$) is in coordination (250,250), whereas the follower AUVs ($AUV_{F1}$) and ($AUV_{F2}$) are in coordination (220,220) and (280,220), respectively, and the heading angle of the AUVs is considered zero. Obstacles are distributed in random positions in the training area. The aim is to train AUVs in such a way that they form a formation with a desired distance of 25 meters and a desired angle of 150 degrees (2.618 rad) and reach the target point safely. Each AUV is equipped with an obstacle avoidance module. The (L) subtitle in all figures denotes the leader, while the ($F_1$) and ($F_2$) subtitles denote the left and right follower, respectively. The training diagrams of all three agents are presented in Fig. 8. The learning curves represent the cumulative reward in each episode as well as the averaged values over all episodes. According to the figure, the average reward gradually increases with training episodes, and the reward value becomes stable.

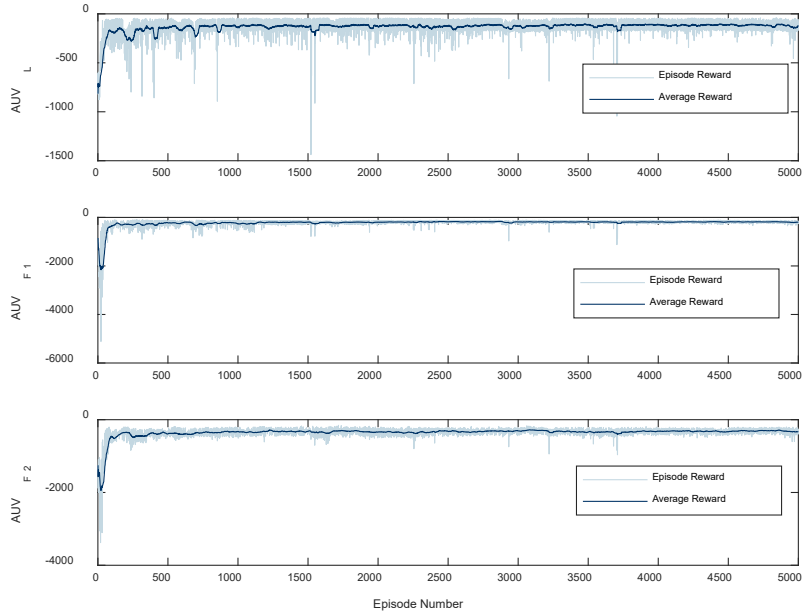

**Fig. 8.** Training result: the first approach rewards and the average reward per episode

### 5.2.1. Obstacle-free formation motion planning results

The leading AUV is responsible for achieving the target, and the following AUVs try to create a predefined formation structure by attaining a specific distance and angle to the leader. Various scenarios for reaching different targets without obstacles have been illustrated in Figs. 9 and 10. For scenario 1, the distance and angle error signals are shown. According to the figures, the desired formation is achieved in different movement scenarios.

### 5.2.2. *Motion planning and formation control in the presence of obstacles*

In this section, the formation motion planning in the presence of obstacles is evaluated. Trajectories of AUVs formation and obstacle avoidance for different scenarios have been illustrated in Fig. 11. In Scenario 1, depending on the position of the target and the obstacles, there is a possibility of colliding with the leader and the left follower. Obstacles are detected by them, and by changing the direction of movement, they prevent the collision with the obstacles, and after passing through the danger zone, they return to the desired formation pattern. In the second scenario, based on the target positions and the obstacles and dimensions of the formation, the followers of AUV reach the target with a slight change in direction. In the third scenario, the obstacle is positioned inside the intended configuration. When the obstacle is revealed to both the leader and the left follower, the left follower bypasses it, and the leader avoids colliding with it by changing direction and then returning to the desired form. Scenario 4 is when the formation faces two obstacles with a distance less than the size of the formation. As can be seen here, both followers have changed direction and returned to the formation path while passing the danger zone. Figures 12 and 13, for instance, show the control signals as well as the system's state variables for scenarios 1 and 2, respectively.

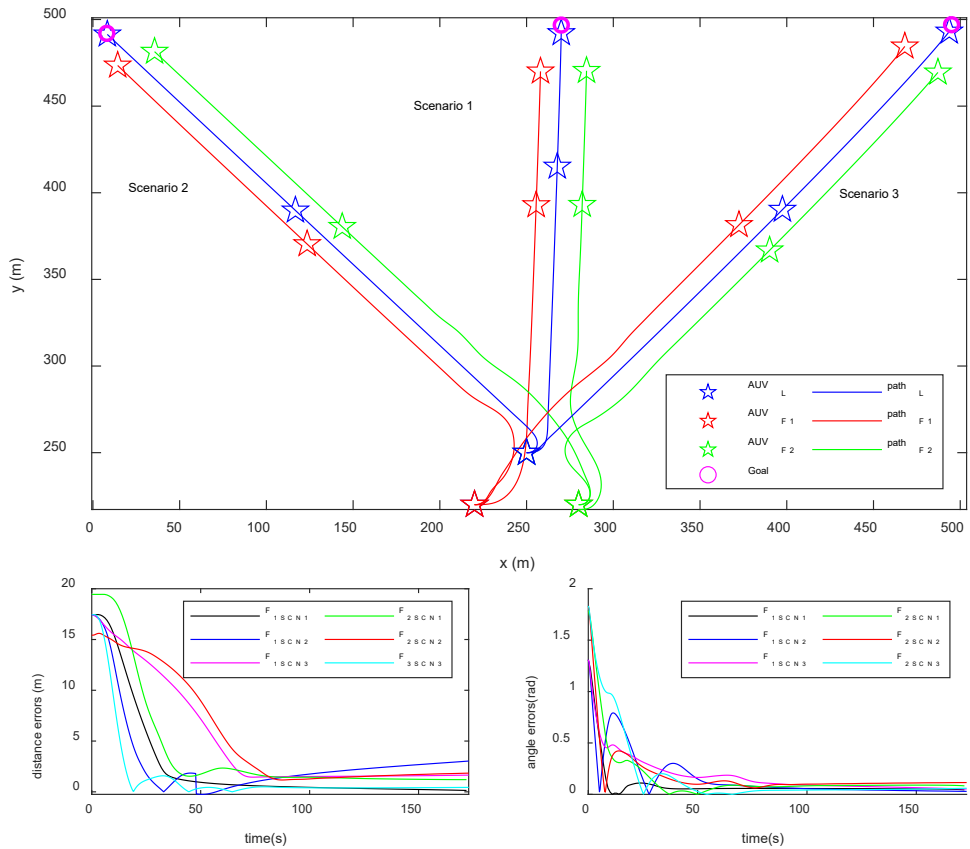

**Fig. 9.** Trajectories of AUVs, distance, and angle formation errors for all obstacle-free scenarios in 5.2.1 (target in the first and second quadrants)

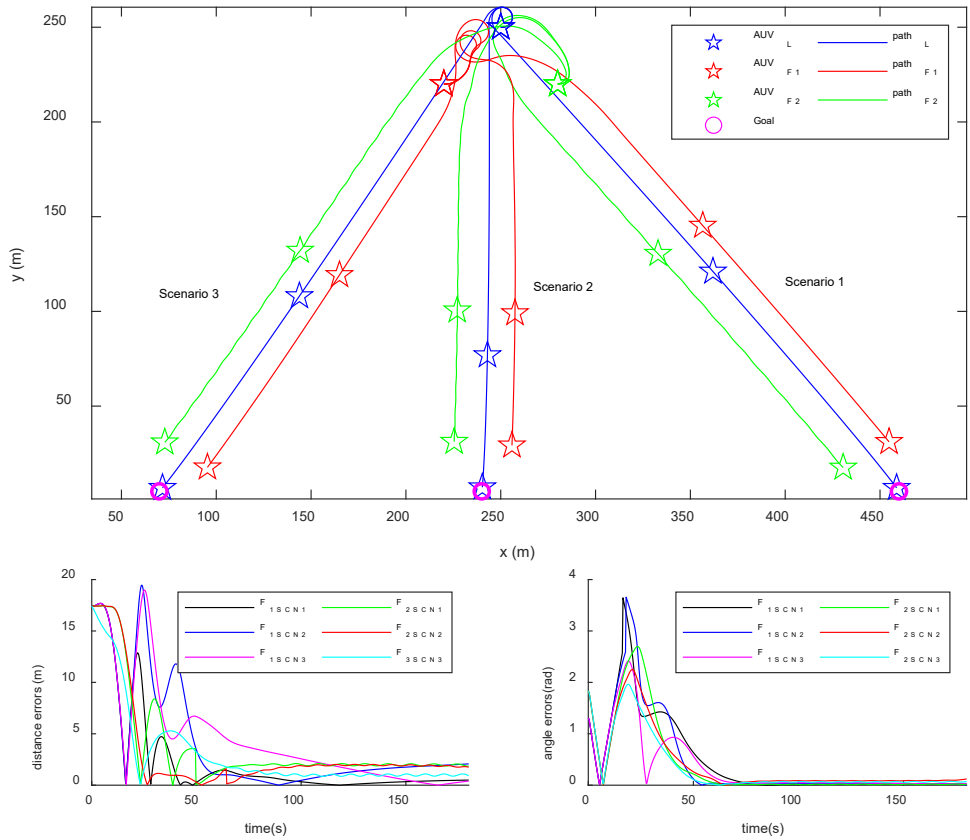

**Fig. 10.** Trajectories of AUVs, distance, and angle formation errors for all obstacle-free scenarios in 5.2.1 (target in the third and fourth quadrants)

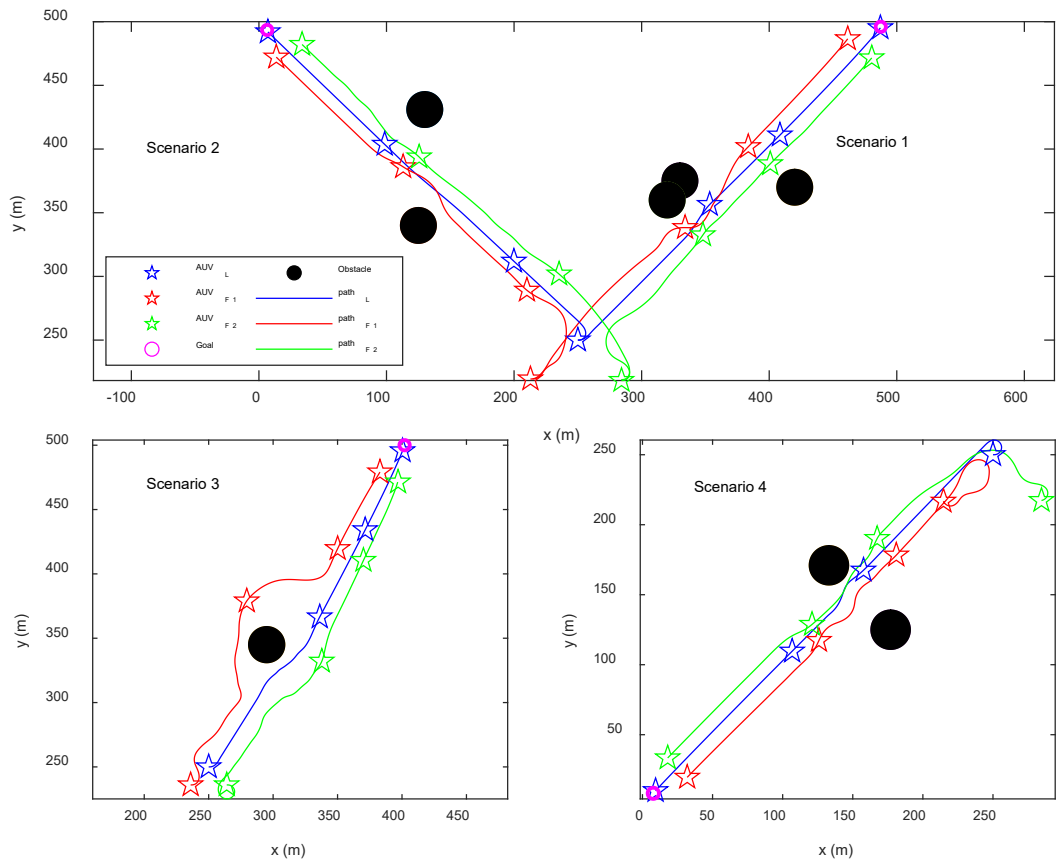

Fig. 11. Trajectories of AUV formation and obstacle avoidance for the first approach

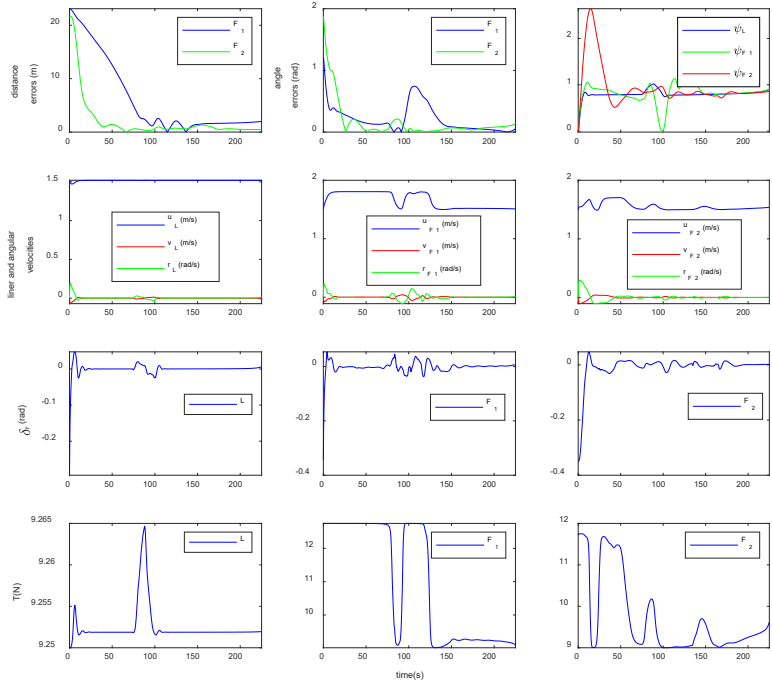

Fig. 12. State variables and control signals of AUVs in scenario 1

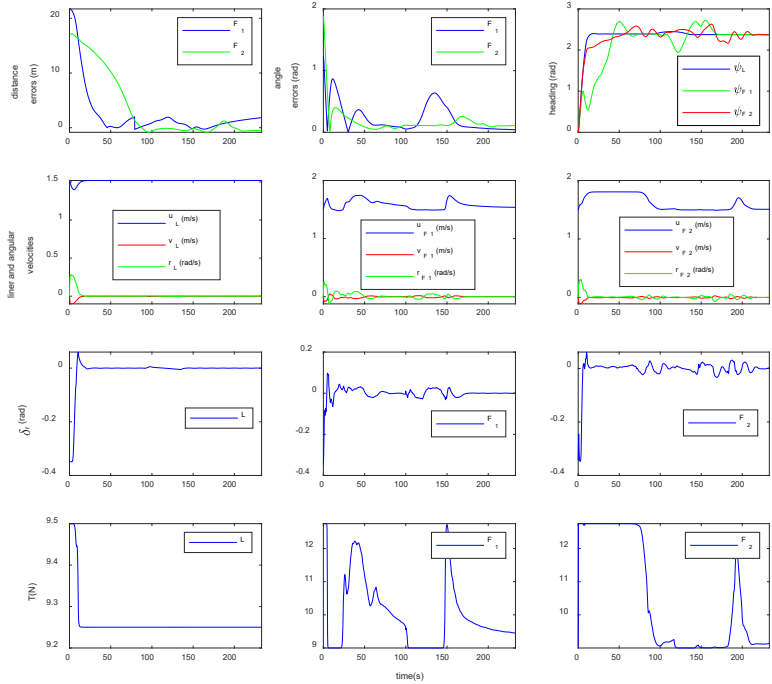

Fig. 13. State variables and control signals of AUVs in scenario 2

## 5.3. Motion planning and formation control: Approach 2

This section presents obstacle avoidance training with the approach of Section 3.2.1. Here, the settings of the environment are the same as in previous sections, and only the leader is equipped with an obstacle avoidance module. In this approach, effective obstacle avoidance is based on the size of the desired formation. Therefore, the training phase begins with the creation of the desired formation, and then obstacles are added to the training phase. The diagrams of network training are presented in Fig. 14. As can be seen in Fig.15, the leader chooses a path by which it can lead the whole group safely towards the destination. Control signals and state spaces for the systems are presented in Fig. 16.

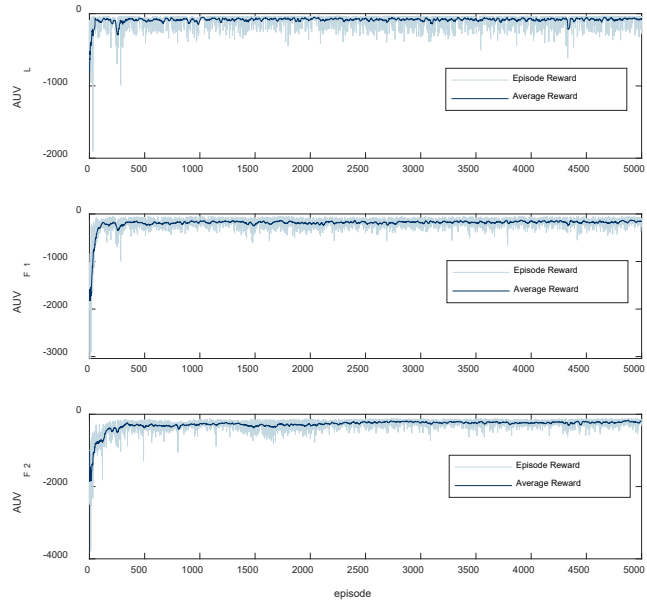

Fig. 14. Training result: the second approach rewards and the average reward per episode

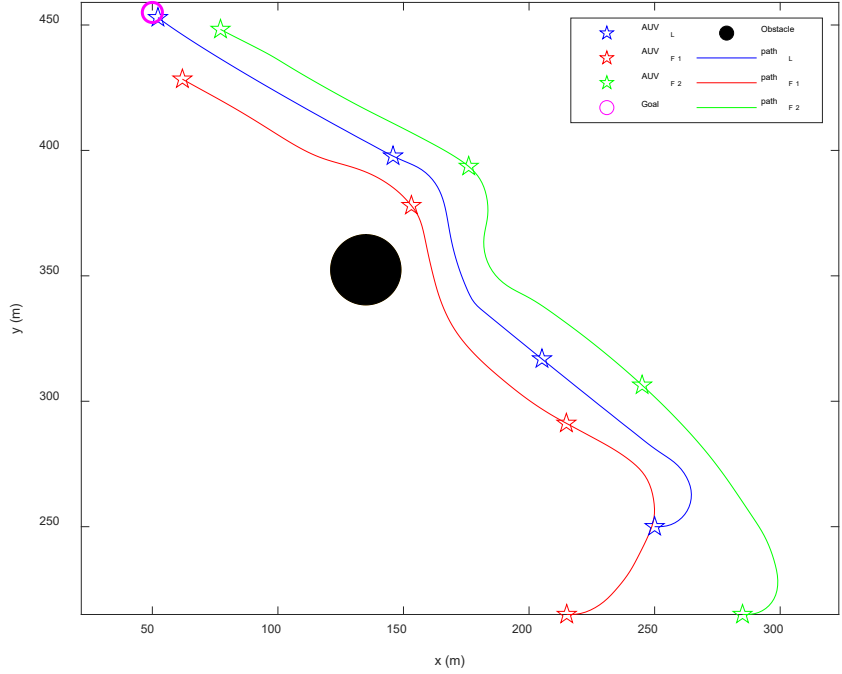

Fig. 15. Trajectories of AUV formation and obstacle avoidance for the second approach

Based on simulations, each obstacle avoidance method performs well in the presence of the obstacle. Both proposed approaches utilize the distributed control architecture. They can be employed, depending on the application.

### 5.4. Simulation in the presence of ocean currents

Ocean currents are horizontal and vertical circulation systems of ocean waters that are generated in various parts of the ocean due to various factors, such as wind friction and gravity. To include ocean currents and how they affect AUV movement, the equations of motion can be written in terms of relative velocity [45]:

$$v_r = v - v_c \qquad (25)$$

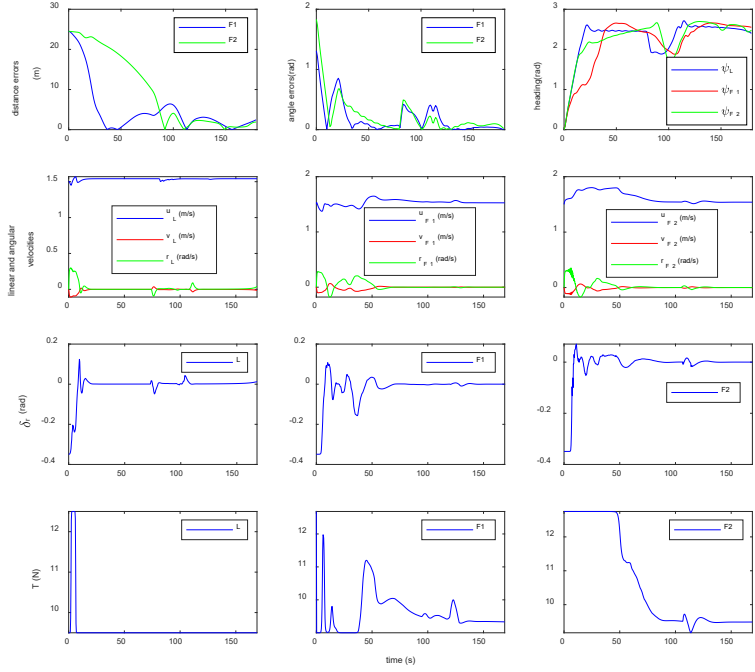

Fig. 16. Error signals, state variables and control signals

where $v_c = [u_c, v_c, 0]^T$ denotes the ocean current in a body-fixed frame, $v = [u, v, r]^T$ represents the linear and angular velocities of AUV in a body-fixed frame. To implement ocean currents, Eq. (1) can be represented in terms of relative velocity as follows [45]:

$$(m - X_{\dot{u}})\dot{u} - my_g\dot{r} - m(v_r r + x_g r^2) = X_{u|u|}|u_r|u_r + X_{vr}v_r r_r + X_{rr}rr + X_{prop} \quad (26)$$

$$(m - Y_{\dot{v}})\dot{v} + (mx_g - Y_{\dot{r}})\dot{r} + m(u_r r_r - y_g r^2) = Y_{v|v|}|v_r|v_r + Y_{r|r|}|r|r + Y_{ur}u_r r +$$
$$Y_{uv}u_r v_r + Y_{uu\delta_r}u_r^2 \delta_r$$

$$(I_z - N_{\dot{r}})\dot{r} + (mx_g - N_{\dot{v}})\dot{v} - my_g\dot{u} + m(x_g u_r r + y_g v_r r) = N_{v|v|}|v_r|v_r + N_{r|r|}|r|r +$$
$$N_{ur}u_r r + N_{uv}u_r v_r + N_{uu\delta_r}u_r^2 \delta_r$$

where $\beta_c$ and $V_c$ are the direction and speed of the ocean current. To evaluate the effect of the ocean current and the system's response to it, different movement directions of AUVs and stochastic ocean current variation are considered. Both formation approaches perform similarly; we just added Approach 1. Accordingly, two distinct scenarios were presented in accordance with Fig. 17, which presents formations designed to reach the destination with and without ocean current. For modeling stochastic ocean currents, a Markov moving average technique is used. The speed and direction of the ocean current change randomly in both scenarios, ranging from 0 to 0.3 m/s and 80 to 140 degrees, respectively. The results indicate that the group can safely reach its destination in formation despite a slight deviation in the AUV's path and formation errors. Consequently, the DRL-based formation motion planning system has satisfactory adaptability and robustness.

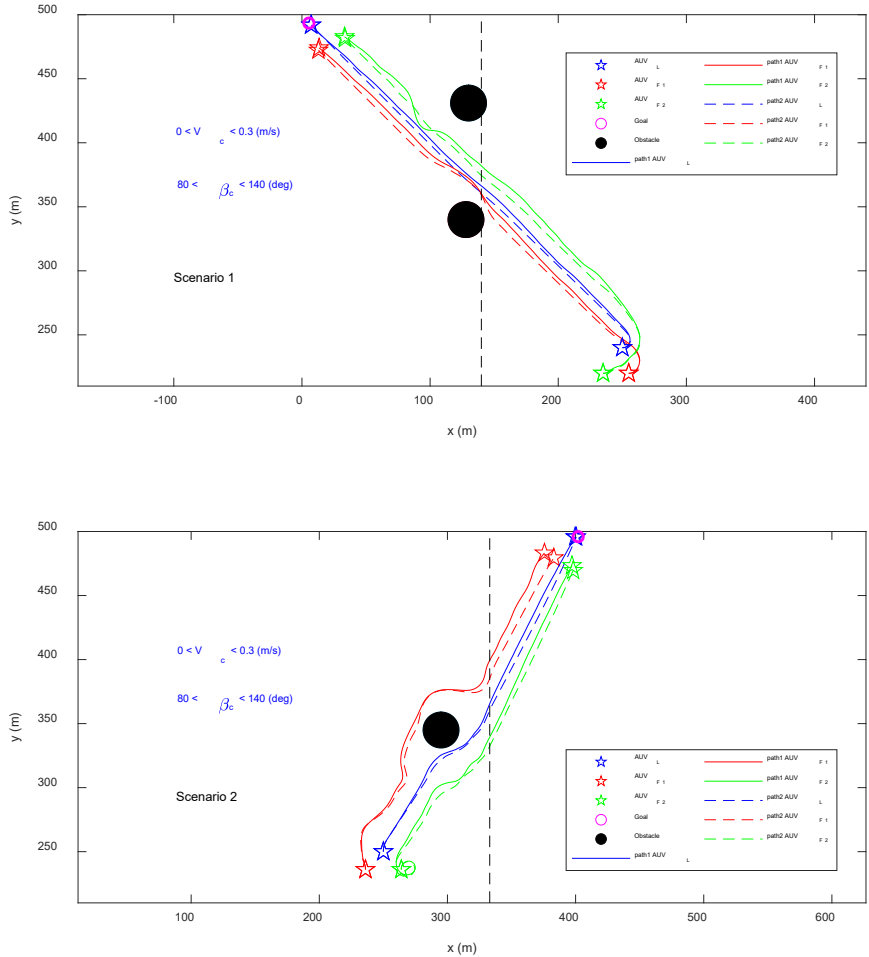

Fig. 17. The trajectories of AUVs in the ocean current for two distinct scenarios, dashed line: AUVs' trajectories without ocean current, solid line: AUVs' trajectories with ocean current

### 5.5. Motion planning and formation control in the presence of communication delay and navigation error

There are other practical challenges like communication delays and navigation (sensing) errors in any AUV formation control systems. Although this paper mainly deals with motion planning and control techniques, those navigation and control errors should be considered somewhere in algorithm design. For this purpose, in this section, the resiliency of the proposed technique is evaluated in the presence of unknown communication delays and navigation errors.

Underwater communication is usually established by acoustic modems which could be a source of errors, mainly variable communication delay [46, 47]. To model underwater communication behavior, a time-varying delay with Rayleigh distribution is considered. The variable delay peak value is considered at 0.1s and it decays after 1.2 s which can be quite challenging to overcome for any control and planning system. AUV navigation is also carried out using a combination of transducers like DVL (Doppler Velocity Log), AHRS (Attitude and heading reference system), depth sensor, etc. which could encounter unknown errors after data fusion. A Markov moving average process is utilized for the modeling of stochastic navigation errors. The navigation error signals that are added to the position and heading states of AUVs are shown in Fig. 18:

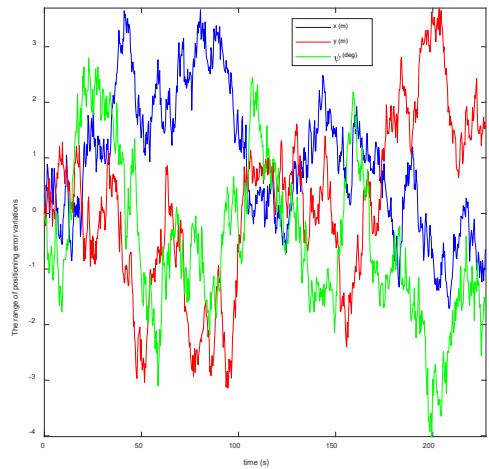

Fig. 18. An example of considered navigation errors for x-y position and $\psi$ angle of AUVs

It should be noted, those communication delays and navigation errors are not considered in the ML training process, therefore they could properly challenge the proposed technique for a perturbed condition. Figure 19 depicts the suggested system's performance in the face of random time-varying delay and random navigation error in the intended scenario. As can be observed from this figure, the proposed algorithm maintains its performance and bounded error in the presence of those perturbations. These results are repeatable for similar scenarios. The obtained performance shows the robustness and generalizability of the proposed intelligent system.

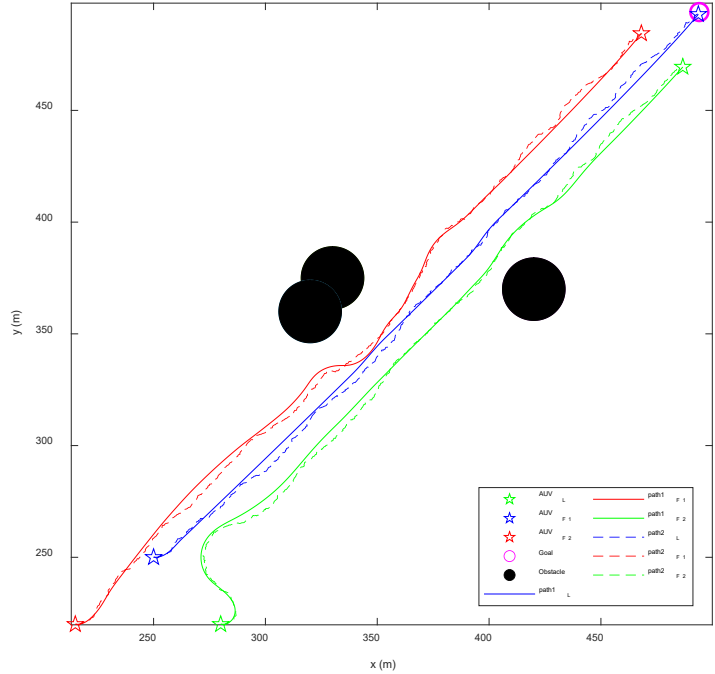

Fig. 19. AUVs' formation paths with navigation and communication errors are shown in dashed lines, while those without these issues are shown in solid lines.

**Implementation Remarks:** Real-time implementation of DRL algorithms is a vibrant and emerging topic. Recent breakthroughs in this field have been led by NVIDIA and other leading AI hardware companies [48-53]. Since the proposed approach deploys two actor and four critic networks (with two hidden layers) for each agent, the implementation can be achieved using comparatively low-cost COTS (commercially off-the-shelf) products like NVIDIA Jetson Nano and TX2 [49, 54, 55]. While successful implementation examples are available in the literature, an exponential surge is likely in the upcoming years [49-52]. For instance, in [51], DRL is employed to control the motion and self-rescue of the x-rudder AUV. Python 2.7 and TensorFlow 1.8.0 are utilized for the training process on a Linux system with 64GB of memory and an Intel i7 7800X processor. The Nvidia Jetson Nano board is employed for implementation purposes. In [49], a hardware and software framework for autonomous, agile quadrotor flight is presented. This platform uses the powerful Nvidia Jetson TX2 board. Zhu et al. [52] have presented a deep reinforcement learning-based end-to-end control and obstacle avoidance framework for visual underwater vehicles. Underwater visual perception and automatic control decision-making are integrated. The model is built with PyTorch and trained on a GeForce GTX 2080 Ti. The BlueROV2 model and Jetson TX2 controller were used for practical testing. One can refer to [49] for a comparison of the available solutions in this area. Hence, the real-time implementation of the proposed algorithms is feasible and envisaged as the next step to complement this research.

## 6. Conclusion

This paper introduces a novel formation control and obstacle avoidance scheme based on deep reinforcement learning. In the suggested system, two solutions for obstacle avoidance are provided through the design of distinct reward functions. In the first approach, safe formation path planning is conducted with all group members equipped with obstacle avoidance sensors. In the second strategy, the leader merely identifies obstacles and guides the entire group toward the target in a safe manner. By designing distinct reward functions for the leader and the follower, given tasks are carried out. The role of the leader is to devise the optimal strategy for moving and guiding the group toward the goal while avoiding potential obstacles. The responsibility for maintaining the desired formation lies with the followers. The performance of the proposed end-to-end motion planning and control approaches is verified through various simulation scenarios. It is shown that the leader and followers are able to maintain their topology during following waypoints and steering clear of obstacles. Moreover, the performance in the presence of ocean currents, communication delays, and sensing errors demonstrates the robustness of the proposed scheme in perturbed realistic conditions. The obtained results verify the merits of the proposed algorithm to be considered for implementation purposes.